\documentclass[letterpaper, 10 pt, conference]{ieeeconf}  % 
\IEEEoverridecommandlockouts                              % This 
\usepackage{graphicx} % for pdf, bitmapped graphics files
\usepackage{amsmath} % assumes amsmath package installed
\usepackage{amssymb}  % assumes amsmath package installed
\usepackage{booktabs}
\usepackage{cite}
\usepackage{lipsum}
\usepackage{comment}
\usepackage{flexisym}
\usepackage{amsmath}
\usepackage{flushend}
\usepackage{algpseudocode}
\usepackage{algorithm}
\usepackage{lipsum}
\usepackage{multirow}
\usepackage{multicol}

\usepackage{svg}

\algnewcommand\algorithmicforeach{\textbf{for each}}
\algdef{S}[FOR]{ForEach}[1]{\algorithmicforeach\ #1\ \algorithmicdo}

\usepackage{subfig}

\usepackage{cite}

\title{Per-frame mAP Prediction for Continuous Performance Monitoring of Object Detection During Deployment} %
\author{Quazi Marufur Rahman, Niko S{\"u}nderhauf and Feras Dayoub %
\thanks{The authors are with the Australian Centre for Robotic Vision at Queensland University of Technology (QUT),
Brisbane, QLD 4001, Australia.  Contact: {\tt\small quazi.rahman@qut.edu.au}}%
}

\begin{document}

\maketitle

\begin{abstract}
Performance monitoring of object detection is crucial for safety-critical applications such as autonomous vehicles that operate under varying and complex environmental conditions. Currently, object detectors are evaluated using summary metrics based on a single dataset that is assumed to be representative of all future deployment conditions. In practice, this assumption does not hold, and the performance fluctuates as a function of the deployment conditions. To address this issue, we propose an introspection approach to performance monitoring during deployment without the need for ground truth data. We do so by predicting when the per-frame mean average precision drops below a critical threshold using the detector's internal features. We quantitatively evaluate and demonstrate our method's ability to reduce risk by trading off making an incorrect decision by raising the alarm and absenting from detection.  
\end{abstract}

\section{Introduction}
Object detection is a crucial part of many safety-critical applications such as robotics and autonomous systems. For safe operation, an autonomous vehicle (AV), for example, needs to accurately locate and identify critical objects like other vehicles and pedestrians on the road. To achieve this goal, there is ongoing research \cite{tian2019fcos, ren2015faster, liu2016ssd, Duan2019CenterNetKT, Bochkovskiy2020YOLOv4OS, He2019RethinkingIP, Cai2018CascadeRD, Li2018DetNetAB, Lin2017FocalLF, Carion2020EndtoEndOD} to improve the speed and accuracy of object detection models. However, due to the discrepancy between training data and deployment environments (i.e., dataset shift~\cite{quionero2009dataset}) and many other unavoidable factors like sensor failure or degraded image quality, a consistent deployment performance can not be guaranteed. Hence, object detection accuracy can fluctuate without any prior notification while deployed on an autonomous vehicle. A silent failure like this in the object detection model is a significant concern. Due to this failure, the AV can cause catastrophic damage if it operates based on erroneous object detection. Undetected performance drops are a significant bottleneck for the widespread deployment of autonomous vehicles in our everyday lives. Hence, for safety, robustness, and reliability, the importance of performance monitoring of a deployed object detection model is paramount.

\begin{figure*}[pt]
\centering
\centering
    \includegraphics[width=0.99\textwidth]{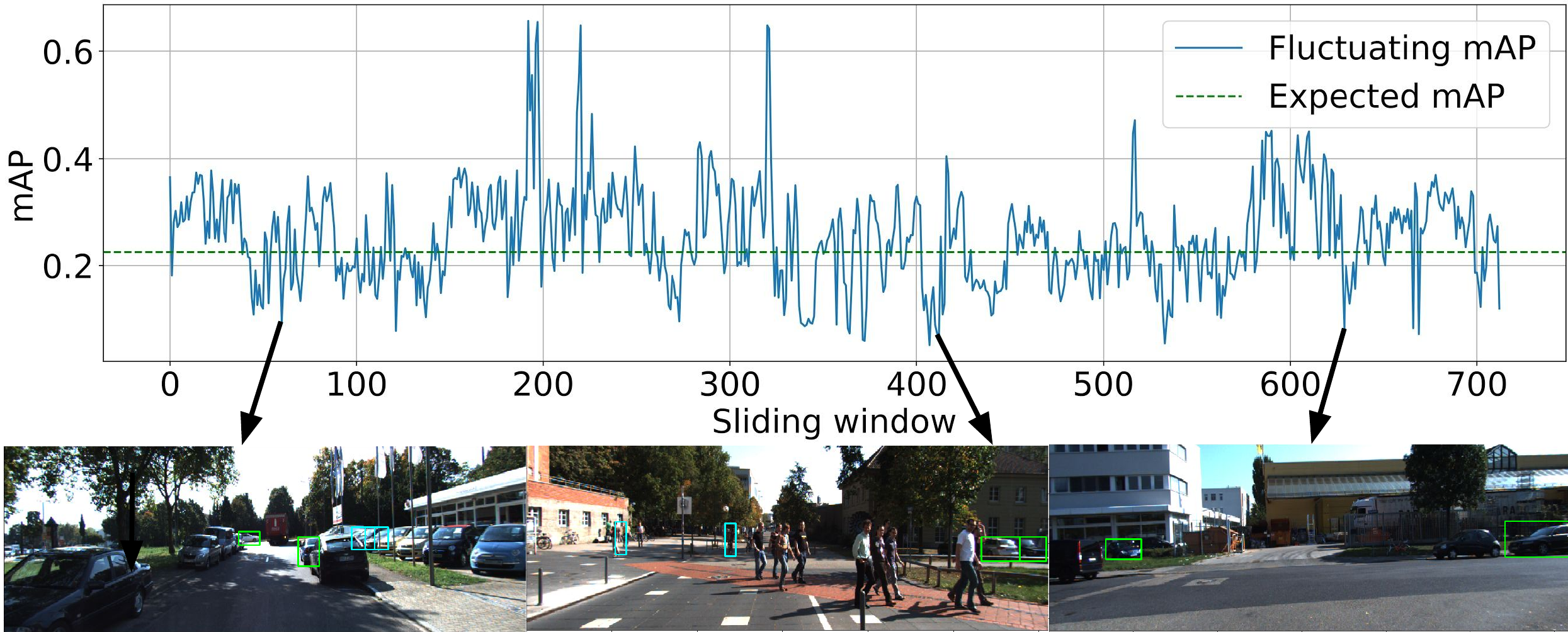}
\caption{First row, an example of object detection performance fluctuation during the deployment, tracked using a sliding window of ten frames. mAP is computed for the ten frames at a time. The dashed line represents a predefined critical threshold. We can see that mAP drops below this threshold from time to time. The second row shows some samples from the low mAP regions. Green and Cyan boxes represent false negative and false positive errors made by the object detector.}
\label{fig:intro fig}
\end{figure*}
The standard practice to prepare an object detection model for deployment is to train and evaluate the model using training and evaluation split of some dataset to measure the accuracy and generalization capacity. Here, the assumption is the training and evaluation data are representative of the real operating environment. However, this assumption does not hold in the context of autonomous vehicles where the operating environment is continuously evolving and might change unexpectedly. Consequently, object detection performance fluctuates without any prior notification. Moreover, the performance might drop below any critical threshold, which can cause a fatal incident. See Figure~\ref{fig:intro fig} for an overview.  

One possible solution is to develop an exceptionally accurate and domain adaptive object detection system for autonomous vehicles. However, it is impossible in most practical circumstances to account for all imaginable future deployment conditions during training. Another approach is to identify when the performance of the deployed object detector drops below a critical threshold. So without the need to increase the detection accuracy directly, a performance drop identifier can protect the autonomous vehicle by providing crucial alerts during periods of silent failure. However, measuring the performance drop directly during deployment is impractical due to the absence of ground-truth data in this phase. Therefore, we advocate equipping object detectors with self-assessment capability to detect instances of performance drop during deployment.

Self-assessment is becoming a prerequisite for any vision-based efficient, safe, and robust robotic system. This capability is often referred to as introspective perception \cite{daftry2016introspective, morris2007robotic}.  For autonomous driving, an introspective object detection system can hand over the control to human drivers when it can predict inconsistency in its operation. There are several works \cite{grimmett2016introspective, Hu2017IntrospectiveEO, Rabiee2019IVOAIV, Mohseni2020PracticalSF} towards addressing the requirement of providing self-assessment in a deep learning based robotic vision system. However, there are very few works towards introspective systems for object detection. To this end, our paper makes the following contributions:
\begin{enumerate}
    \item We propose an introspective approach to performance monitoring of object detection during deployment without access to ground truth labels.

    \item We propose an internal integrated feature based on the mean, max and statistics pooling techniques for performance monitoring. 

    \item We introduce the use of per-frame mAP prediction for continuous performance monitoring of object detectors. 
\end{enumerate}

The rest of the paper is organized as follows: In Section~\ref{sec:literature review}, we review the related works on introspective perception systems. In Section~\ref{sec:approach overview}, we introduce our framework to find the performance drop for an object detection system. Section~\ref{sec:experimental setup} outlines our experimental setup. Section~\ref{sec:Evaluation and Results} presents the results and finally in Section~\ref{sec:conclusion} we draw conclusions and suggest areas for future work.

\section{Related Work}
\label{sec:literature review}
In robotics,  the idea of self-assessment was introduced by \cite{morris2007robotic} to achieve reliable performance in a real environment. They described this self-assessment as the introspection capability of a mobile robot while operating in an unknown environment.  Later \cite{grimmett2016introspective} and \cite{Triebel2013DrivenLF} adopted the idea of introspection for classification and semantic mapping respectively in the context of robotics. These works examine the output of the underlying system to predict the likelihood of failure on any given input. 

Another line of research is to predict the perception system performance from the input itself. In this paradigm, \cite{Jammalamadaka2012HasMA} introduced an evaluator algorithm to predict the failure of a human pose estimation model. Zhang \textit{et al.}~\cite{zhang2014predicting} introduced the terminology \textit{basesys} and \textit{alert} in failure prediction context. They proposed a general framework where \textit{alert} is used to raise a warning when the underlying system \textit{basesys} fails to make a correct decision from an input. Daftry \textit{et al.}~\cite{daftry2016introspective} proposed an introspective framework to predict an image classifier model failure deployed on a micro aerial vehicle. Following a similar methodology, \cite{wang2018towards} proposed a model to predict how hard an image is for an underlying classifier. Using a probabilistic model, \cite{Gurau2018LearnFE} predicted the probable performance of a robot's perception system based on past experience in the same location.

Recently, confidence estimation and Bayesian approaches for uncertainty estimation have gained popularity to detect how well the underlying model has performed on the input. TrustScore \cite{Jiang2018ToTO}, Maximum Class Probability \cite{Hendrycks2017ABF} and  True Class Probability \cite{corbiere2019addressing} are some of the works that measure the confidence of the underlying model for a given task using the confidence estimation paradigm. Using a Bayesian approach, \cite{Gal2016DropoutAA} proposed to use dropout as a Bayesian approximation technique to represent model uncertainty. Following their work, \cite{Devries2018LeveragingUE, Huang2018EfficientUE} have used dropout sampling to identify the quality of image and video segmentation network. Here, all of these works focus on predicting model failure using different approaches for classification and segmentation tasks.

In the context of object detection, \cite{Miller2018DropoutSF, Cheng2018DecoupledCR} have used different approaches to identify the failure of an object detection system. Both of these works are beneficial to identify false positive errors made by an object detector. Whereas, \cite{Rahman2019DidYM} and \cite{Ramanagopal2018FailingTL} have proposed different approaches to detect false negative errors made by an object detection model. Our proposed approach differs from these methods in that we can detect images with low per-frame mAP, which covers both false positive and false negative errors as well as poor object localization.

\section{Approach Overview}
\label{sec:approach overview}

In this section we describe our proposed framework to predict the performance drop of an object detection system during deployment without using any ground-truth data. We assume that the deployed object detection model weight remains frozen during this phase. First of all, we will define the problem.

Assuming we have an object detector $O$ with backbone deep neural network $B$, $O$ is trained to detect a set of objects $T$ from a training dataset, $D_{t}$. We also have an evaluation dataset $D_{e}$, similarly distributed as $D_{t}$. $D_{e}$ contains a set of images $I = \{I_{1}, I_{2}\dots I_{n}\}$ and corresponding annotations $L = \{L_{1}, L_{2}\dots L_{n}\}$ per image. After the object detection training phase, $O$ is applied on $D_{e}$ to detect all the objects from $T$. Thus, we get a set of predictions per image, $P = \{P_{1}, P_{2}\dots P_{n}\}$. Using the pairs of annotations and predictions per image $(L_{i}, P_{i})$, we compute the per-frame mAP, $M = \{M_{1}, M_{2}\dots M_{n}\}$ following the procedure at \cite{lin2014microsoft}. Here, per-frame mAP quantifies $O$’s performance to detect all the existing objects in each image.

We assign each image of $D_{e}$ into \textit{success} and \textit{failure} classes using the Equation~\ref{eqn:failure success threshold}. Here $\lambda$ is chosen to be the $k^{th}$ percentile of $M$.  The \textit{failure} class contains the $k\%$ image frames from the $D_{e}$ where $O$ was not accurate enough to detect the available objects. The choice of $k$ here is application specific. We want to train the introspective perception system \textit{alert} to predict the images similar to the \textit{failure} class where per-frame mAP will be lower than $\lambda$.

\begin{equation}
    \mathcal{L}(I)= 
\begin{cases}
    failure, & mAP_{\texttt{per-frame}} < \lambda\\
    success, & \text{otherwise}
\end{cases}
\label{eqn:failure success threshold}
\end{equation}

To train the \textit{alert} we use features $F = \{F_{1}, F_{2}\dots F_{n}\}$ for each image from $D_{e}$. Following the failure prediction network proposed by \cite{wang2018towards} and \cite{corbiere2019addressing}, the final convolutional layer of backbone $B$ is used to extract all the necessary features. Assuming that, there are $N$ channels at the last layer of $B$ and each activation map is of size $W \times H$, we apply Equation~\ref{eqn:avg_pooling_eqn} on the last layer to extract the mean pooling feature $F_{mean}$. Here, $f(x,y)$ represents the spatial unit of each activation map.
\begin{equation}
    F_{mean} = \frac{\sum_{x=1}^{H}\sum_{y=1}^{W}f(x, y)}{W \cdot H}
    \label{eqn:avg_pooling_eqn}
\end{equation}

\noindent Applying Equation~\ref{eqn:max_pooling_eqn} on the last layer of $B$, we generate the max pooling feature $F_{max}$.

\begin{equation}
    F_{max} = \max_{x\in[1, H]}\max_{y\in[1, W]}f(x, y)
    \label{eqn:max_pooling_eqn}
\end{equation}

\noindent Inspired by \cite{snyder2017deep}, we calculate the standard deviation from each activation map to generate the statistics pooling feature $F_{std}$ following the Equation~\ref{eqn:std_pooling_eqn}. Here $std(f_{i})$ calculates the standard deviation of $i^{th}$ feature map.

\begin{equation}
    F_{std} = std(f_{1}) \oplus std(f_{2}) \oplus \dots std(f_{N})
    \label{eqn:std_pooling_eqn}
\end{equation}

\noindent All the features described above are concatenated together to generate the feature $F_{mean\_max\_std}$ for the \textit{alert} system. Equation~\ref{eqn:hybrid feature} formulates this process.

\begin{equation}
    F_{mean\_max\_std} = F_{mean} \oplus F_{max} \oplus F_{std}
    \label{eqn:hybrid feature}
\end{equation}

\begin{figure}[t]
\centering
\centering
    \includegraphics[width=0.99\columnwidth]{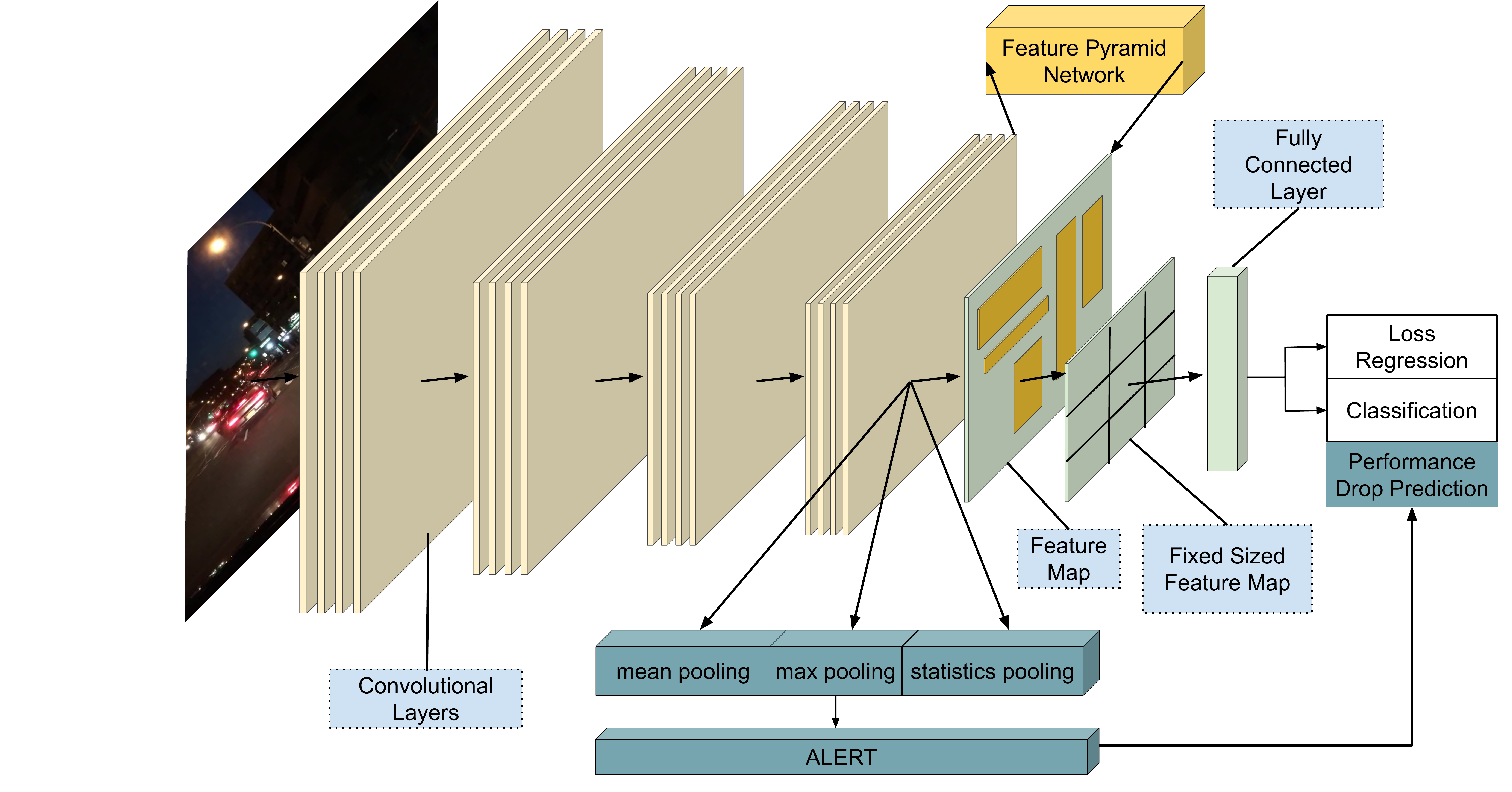}
\caption{The architecture of our proposed \textit{basesys} and \textit{alert} system together. The last convolutional feature of the backbone is pooled using the mean, max and statistical pooling layer to generate the feature for \textit{alert}. \textit{Alert} consists of a binary classifier that predict the performance drop of the \textit{basesys}.}
\label{fig:architecture}
\end{figure}

We train a binary classifier using the $F_{mean\_max\_std}$ feature and the corresponding labels from Equation~\ref{eqn:hybrid feature} and Equation~\ref{eqn:failure success threshold} respectively. The classifier is trained to predict the probability of an image feature to be in the \textit{failure} class. Following \cite{zhang2014predicting}, we will refer to the object detection model and its corresponding binary classifier as \textit{basesys} and \textit{alert} respectively. Figure~\ref{fig:architecture} shows the incorporation between the \textit{basesys} and \textit{alert} system.

\section{Experimental Setup}
\label{sec:experimental setup}
In this section we will describe the settings that we have used to train the \textit{basesys} and \textit{alert} system.

\paragraph{\textbf{Datasets:}} We used all  images from \textit{kitti} \cite{KITTI} dataset and one frame per video from \textit{bdd} \cite{yu2018bdd100k} dataset to train both \textit{basesys} and corresponding \textit{alert} system. Randomly selected $60\%$, $20\%$ and $20\%$ images from both dataset have been used for \textit{basesys} training, evaluation and testing purpose.  As person and car classes are available in both of these dataset, we used these two objects as \textit{basesys} target class. After the object detection training, \textit{basesys} is used to detect person and car from the $20\%$ evaluation split. Based on \textit{basesys} performance on the evaluation split, we collect image features and labels for the \textit{alert} system following the procedure described in Section~\ref{sec:approach overview}. Thus the features and labels collected from \textit{basesys} evaluation split works as training dataset for the corresponding \textit{alert} system. To test the \textit{alert}, we first apply \textit{basesys} on the testing split and measure its per-frame performance drop, which works as the testing data for the \textit{alert} system. In some of our experiments, we will train and test \textit{basesys} and \textit{alert} using training and testing split of a single dataset. We will refer this settings as \textit{similar dataset}. In rest of the experiments, \textit{basesys} and \textit{alert} will be trained using training split of one dataset and tested using testing split of another dataset. This arrangements will be referred as \textit{cross dataset} settings.

\begin{table}[h]
\centering
\caption{\textit{Basesys} mean average precision (mAP) using ResNet18 and ResNet50 backbone. Here \textit{basesys} is trained and tested using \textit{similar dataset} and \textit{cross dataset} settings. \textit{Basesys} accuracy drops when trained and tested on different dataset.}
\label{tab:all_map}
\begin{tabular}{|c|c|c|c|c|c|c|c|}
\hline
\multicolumn{2}{|c|}{\multirow{2}{*}{ResNet18}} &
  \multicolumn{2}{c|}{testing dataset} &
  \multicolumn{2}{c|}{\multirow{2}{*}{ResNet50}} &
  \multicolumn{2}{c|}{testing dataset} \\ \cline{3-4} \cline{7-8} 
\multicolumn{2}{|c|}{} & kitti & bdd   & \multicolumn{2}{c|}{} & kitti & bdd   \\ \hline
\multirow{2}{*}{\begin{tabular}[c]{@{}c@{}}training\\ dataset\end{tabular}} &
  kitti &
  0.292 &
  0.130 &
  \multirow{2}{*}{\begin{tabular}[c]{@{}c@{}}training\\ dataset\end{tabular}} &
  kitti &
  0.377 &
  0.182 \\ \cline{2-4} \cline{6-8} 
         & bdd         & 0.200 & 0.331 &          & bdd        & 0.259 & 0.499 \\ \hline
\end{tabular}
\end{table}
\begin{table*}[t]
	\centering
	\caption{Area Under the Precision Recall Curve (AUPRC) and Area Under the ROC Curve (AUROC) score for \textit{alert} system in the similar dataset settings. Here \textit{alert} is used to identify \textit{basesys} performance drop in a known environment. The notation \textit{A/B/C} denotes that \textit{basesys} and \textit{alert} is trained on dataset \textit{A} using backbone \textit{C} and \textit{alert} is used to identify \textit{basesys} performance drop on dataset\textit{B}.}
	\label{tab:same_dataset_auprc_auroc}
	\begin{tabular}{|c|c|c|c|c|c|c|c|c|}
		\hline
		\multicolumn{1}{|l|}{} & \multicolumn{2}{c|}{\textit{kitti/kitti/18}}     & \multicolumn{2}{c|}{\textit{kitti/kitti/50}} & \multicolumn{2}{c|}{\textit{bdd/bdd/18}}     & \multicolumn{2}{c|}{\textit{bdd/bdd/50}} \\ \hline
		Feature                 & AUPRC       & AUROC      & AUPRC        & AUROC & AUPRC       & AUROC      & AUPRC        & AUROC \\ \hline
		n\_proposals       & 0.180 & 0.128 & 0.186 & 0.110          & 0.205 & 0.368 & 0.197 & 0.363          \\ \hline
		mean\_conf\_score   & 0.205 & 0.320 & 0.192 & 0.358          & 0.452 & 0.653 & 0.463 & 0.665          \\ \hline
		classifier & 0.728 & 0.851 & 0.689 & 0.831          & 0.498 & 0.744 & 0.488 & 0.734          \\ \hline
		places365  & 0.654 & 0.799 & 0.670 & 0.823          & 0.516 & 0.753 & 0.507 & 0.744          \\ \hline
		layer      & 0.760 & 0.890 & 0.480 & 0.764          & 0.622 & 0.814 & 0.587 & \textbf{0.798} \\ \hline
		mean       & 0.738 & 0.876 & 0.602 & 0.822          & 0.587 & 0.800 & 0.549 & 0.777          \\ \hline
		max        & 0.756 & 0.887 & 0.673 & 0.819          & 0.621 & 0.811 & 0.587 & 0.790          \\ \hline
		mean\_std  & 0.747 & 0.879 & 0.689 & \textbf{0.855} & 0.609 & 0.815 & 0.577 & 0.791          \\ \hline
		mean\_max  & 0.777 & 0.898 & 0.708 & 0.841          & 0.627 & 0.818 & 0.587 & 0.793          \\ \hline
		mean\_max\_std         & \textbf{0.781} & \textbf{0.902} & \textbf{0.712}  & 0.846     & \textbf{0.633} & \textbf{0.820} & \textbf{0.595}  & 0.795     \\ \hline
	\end{tabular}
\end{table*}

\begin{table*}[t]
	\centering
	\caption{Area Under the Precision Recall Curve (AUPRC) and Area Under the ROC Curve (AUROC) for \textit{alert} in the cross dataset settings. Here \textit{alert} is identifying \textit{basesys} performance drop in an unknown environment. The notation \textit{A/B/C} denotes that \textit{basesys} and \textit{alert} is trained on dataset \textit{A} using backbone \textit{C} and \textit{alert} is used to identify \textit{basesys} performance drop on dataset \textit{B}.}
	\label{tab:cross_dataset_auprc_auroc}
	\begin{tabular}{|c|c|c|c|c|c|c|c|c|}
		\hline
		& \multicolumn{2}{c|}{\textit{bdd/kitti/18}}     & \multicolumn{2}{c|}{\textit{bdd/kitti/50}}     & \multicolumn{2}{c|}{\textit{kitti/bdd/18}}     & \multicolumn{2}{c|}{\textit{kitti/bdd/50}}     \\ \hline
		Feature         & AUPRC       & AUROC      & AUPRC       & AUROC      & AUPRC       & AUROC      & AUPRC       & AUROC      \\ \hline
		n\_proposals       & 0.558 & 0.381 & 0.736 & 0.447 & 0.531 & 0.465 & 0.538 & 0.497 \\ \hline
		mean\_conf\_score   & 0.641 & 0.507 & 0.786 & 0.566 & 0.675 & 0.623 & 0.730 & 0.710 \\ \hline
		classifier & 0.783 & 0.629 & 0.818 & 0.483 & 0.672 & 0.663 & 0.557 & 0.578 \\ \hline
		places365  & 0.781 & 0.624 & 0.821 & 0.493 & 0.857 & 0.837 & 0.630 & 0.637 \\ \hline
		layer      & 0.780 & 0.684 & 0.815 & 0.580 & 0.868 & 0.836 & 0.662 & 0.670 \\ \hline
		mean       & 0.754 & 0.665 & 0.809 & 0.563 & 0.858 & 0.831 & 0.733 & 0.753 \\ \hline
		max        & 0.778 & 0.682 & 0.813 & 0.582 & 0.647 & 0.605 & 0.661 & 0.686 \\ \hline
		mean\_std  & 0.759 & 0.672 & 0.815 & 0.569 & 0.855 & 0.823 & 0.751 & 0.768 \\ \hline
		mean\_max  & 0.786 & 0.692 & 0.822 & 0.586 & 0.826 & 0.786 & 0.701 & 0.726 \\ \hline
		mean\_max\_std & \textbf{0.790} & \textbf{0.696} & \textbf{0.822} & \textbf{0.587} & \textbf{0.883} & \textbf{0.856} & \textbf{0.833} & \textbf{0.825} \\ \hline
	\end{tabular}
\end{table*}

\paragraph{\textbf{Basesys Training:}} We have used Faster RCNN object detection network \cite{ren2015faster} as the \textit{basesys} in all of our experiments. \textit{Basesys} has been trained using transfer learning to detect person and car object from both \textit{kitti} and \textit{bdd} dataset. Two different versions of Residual Neural Network \cite{he2016deep}, ResNet18 and ResNet50 have been used as the \textit{basesys} backbone. In our experiments, the \textit{basesys}, trained using RestNet50 backbone has performed better than the ResNet18 backbone. Table~\ref{tab:all_map} shows comparative performance using the mean average precision (mAP) for all different \textit{basesys} and dataset combinations.

\paragraph{\textbf{Feature Collection:}} We experimented with multiple features to find the most suitable one for the proposed \textit{alert} system. The first set of features are collected from \textit{basesys} bounding box proposals.
\begin{itemize}
    \item \textit{mean\_conf\_score}: This feature exploits object proposal confidence score to determine  \textit{basesys} performance drop. As \textit{basesys} proposes multiple bounding boxes with corresponding confidence scores and labels during object detection, we use the mean of confidence scores which are greater than $0.5$ to build the first feature. Here, a lower mean confidence score indicates a potential performance drop in the \textit{basesys}. 
    \item \textit{n\_proposals}: We assume that a crowded environment might be a factor for \textit{basesys} performance drop. To evaluate this assumption, we used the number of proposals having a confidence score greater than $0.5$ as a performance drop indicator.
\end{itemize}
The second set of features are collected from two external deep convolutional neural networks.
\begin{itemize}
\item \textit{classifier:} Two different versions of Residual neural network, Resnet18 and ResNet50 have been used to extract image features to train the \textit{alert} system. Both of these networks are pre-trained on ImageNet \cite{deng2009imagenet} dataset.
\item \textit{places365:} We used ResNet18 and ResNet50 network pre-trained on Places365 \cite{Zhou2018PlacesA1} dataset to extract features to train the \textit{alert} system. 

\begin{figure*}[pht]
	\centering
	\centering
	\includegraphics[width=0.99\textwidth]{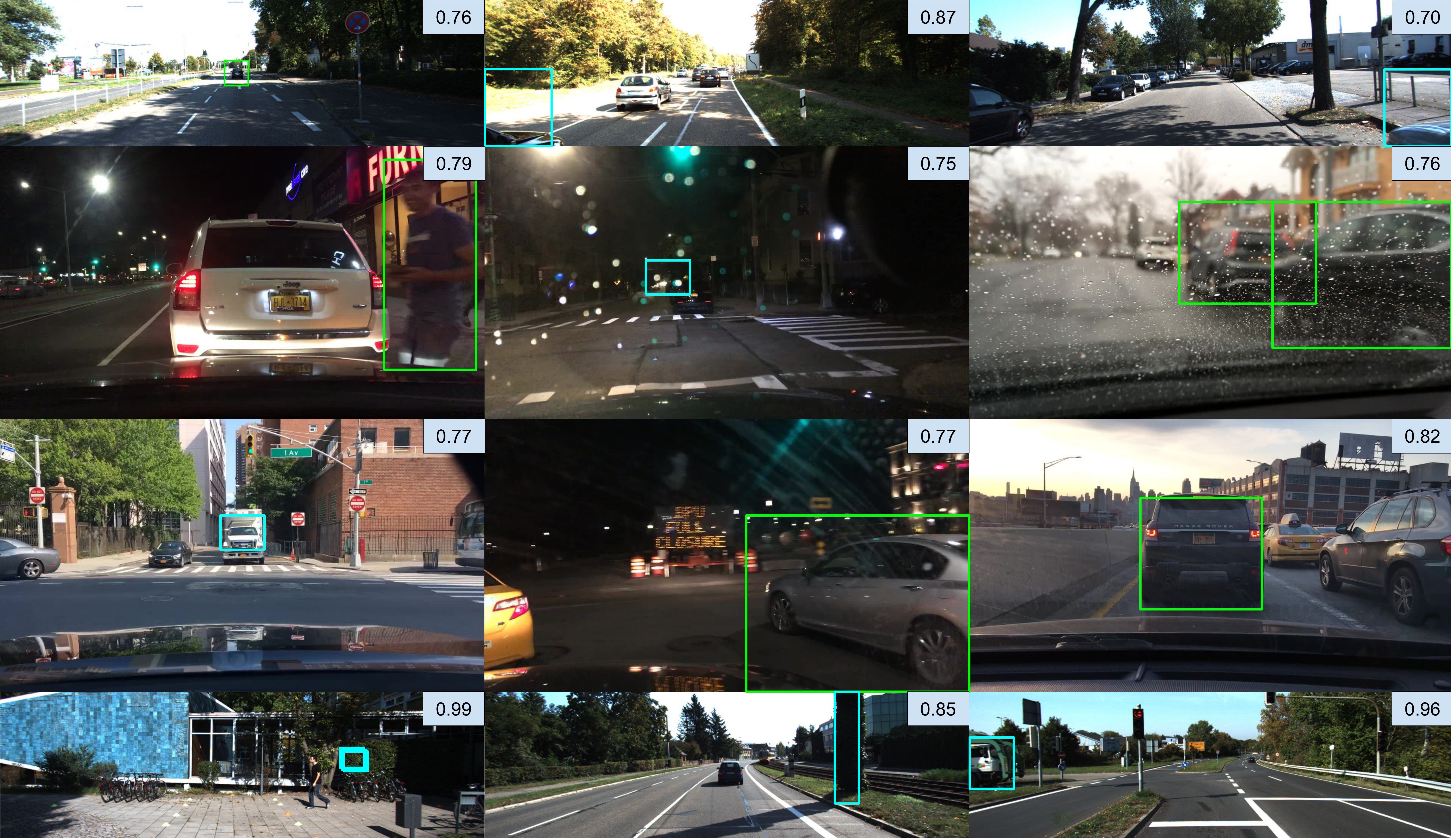}
	\caption{Examples of \textit{alert} prediction to identify \textit{basesys} performance drop. Here the Green and Cyan bounding boxes show the false negative and false positive errors respectively made by an object detector. \textit{Alert} prediction is showed at the upper right corner of each image. The first row shows samples from the \textit{kitti/kitti/50} experimental settings. The second, third and fourth row show samples from the \textit{bdd/bdd/18}, \textit{kitti/bdd/18} and \textit{bdd/kitti/50} experiments.}
	\label{fig:samples}
\end{figure*}
In both cases, average pooling has been used at the final convolutional layer to extract the necessary image features. 
\end{itemize}
We use the  \textit{basesys} backbone to extract the third set of features. These will be referred as the internal features.
\begin{itemize}
\item \textit{layer:} We applied the mean-pooling operation in all of the convolutional layers of the backbone and concatenated them to create this feature.
\item \textit{mean}, \textit{max} and \textit{std}: Applying the mean, max and statistics pooling technique described in Section~\ref{sec:approach overview} at the last convolutional layer of \textit{basesys} backbone, we extracted the \textit{mean}, \textit{max} and \textit{std} features.
\item \textit{mean\_std} and \textit{mean\_max:} Using the concatenation operation and following the feature generation technique proposed in \cite{snyder2017deep} and \cite{Zhang2018DelvingIF}, we generate two new features \textit{mean\_std} and \textit{mean\_max} using the \textit{mean}, \textit{max} and \textit{std} feature.
\item \textit{mean\_max\_std:} This feature is generated by applying the Equation~\ref{eqn:hybrid feature} at the last convolutional layer of \textit{basesys} backbone.
\end{itemize}

\paragraph{\textbf{\textit{Alert} Training:}} We used a multi layer fully connected binary classifier with $50\%$ dropout rate to train all the \textit{alert} systems. Besides, we used binary cross entropy loss with balanced sampling to train the \textit{alert} network.

\section{Evaluation and Results }
\label{sec:Evaluation and Results}

\subsection{AUPRC and AUROC Metrics}
This section summarizes the \textit{alert} accuracy using Area Under the Precision Recall Curve (AUPRC)  and Area Under the ROC Curve (AUROC)  metric. Here, we will refer all our experimental settings using the notation \textit{A/B/C}. It means the \textit{basesys} and \textit{alert} are trained on dataset \textit{A} using backbone \textit{C} and \textit{alert} is used to identify \textit{basesys} performance drop on dataset \textit{B}. Here \textit{C} can be $18$ or $50$, resembling the ResNet18 and ResNet50 backbone for the \textit{basesys}.

Table~\ref{tab:same_dataset_auprc_auroc} summarizes the \textit{alert} accuracy for similar dataset settings. Our proposed \textit{mean\_max\_std} feature achieves $0.781$ and $0.902$ as AUPRC and AUROC score, and outperforms all other features in the case of \textit{kitti/kitti/18}. For \textit{kitti/kitti/50}, \textit{bdd/bdd/18} and \textit{bdd/bdd/50} experimental settings, features collected from the \textit{basesys} performs better than all other features in terms of AUPRC and AUROC score.

The proposed \textit{alert} system is beneficial for cross dataset settings too. Table~\ref{tab:cross_dataset_auprc_auroc} shows the AUPRC and AUROC scores for \textit{alert} when it is used to identify \textit{basesys} performance when deployed on an unknown environment. For \textit{bdd/kitti/18} settings, \textit{alert} achieves $0.790$ and $0.696$ as AUPRC and AUROC score respectively when used with the \textit{mean\_max\_std} feature. In all cross dataset experimental settings, \textit{mean\_max\_std} features outperforms all other features for identifying \textit{basesys} performance drop.

\begin{table}[h]
\centering
\caption{The true warning rate of the \textit{alert} system to identify \textit{basesys} performance drop.}
\label{tab:warning rate}
\begin{tabular}{|c|c|c|c|c|c|c|c|}
\hline
\multicolumn{2}{|c|}{\multirow{2}{*}{ResNet18}} &
  \multicolumn{2}{c|}{testing dataset} &
  \multicolumn{2}{c|}{\multirow{2}{*}{ResNet50}} &
  \multicolumn{2}{c|}{testing dataset} \\ \cline{3-4} \cline{7-8} 
\multicolumn{2}{|c|}{} & kitti & bdd   & \multicolumn{2}{c|}{} & kitti & bdd   \\ \hline
\multirow{2}{*}{\begin{tabular}[c]{@{}c@{}}training\\ dataset\end{tabular}} &
  kitti &
  59.1\% &
  81.8\% &
  \multirow{2}{*}{\begin{tabular}[c]{@{}c@{}}training\\ dataset\end{tabular}} &
  kitti &
  66.0\% &
  78.7\% \\ \cline{2-4} \cline{6-8} 
         & bdd         & 79.3\% & 55.4\% &          & bdd        & 81.4\% & 52.4\% \\ \hline
\end{tabular}
\end{table}
\subsection{True Warning Rate}
Using the best performing feature, \textit{mean\_max\_std}, we use the true warning rate metric to determine the quality of the \textit{alert} system in raising a warning against \textit{basesys} performance drop. Here, warning rate is the ratio of correctly raised warning vs the total number of frames with per-frame mAP below the critical threshold. Table~\ref{tab:warning rate} shows the true warning rate raised by \textit{alert} system.

The results in Table~\ref{tab:warning rate} show that in \textit{cross dataset} settings the true warning rate is higher than the \textit{similar dataset} settings. As \textit{basesys} accuracy drops in cross dataset settings (Table~\ref{tab:all_map}), \textit{alert} becomes more useful in these cases by identifying the critical cases. When the detector with ResNet50 backbone is trained on BDD and tested on Kitti, \textit{alert} can identify $81.4\%$ of the frames where \textit{basesys} per-frame mAP is lower than the critical threshold. 
Figure~\ref{fig:samples} displays multiple samples of \textit{alert} raising the alarm and flagging frames where \textit{basesys} performance drop below a critical threshold of $0.5$. The frames show conditions such as night, rain, cluttered environments.    

\begin{figure}[]
\centering
\subfloat[][]{\includegraphics[width=0.49\columnwidth]{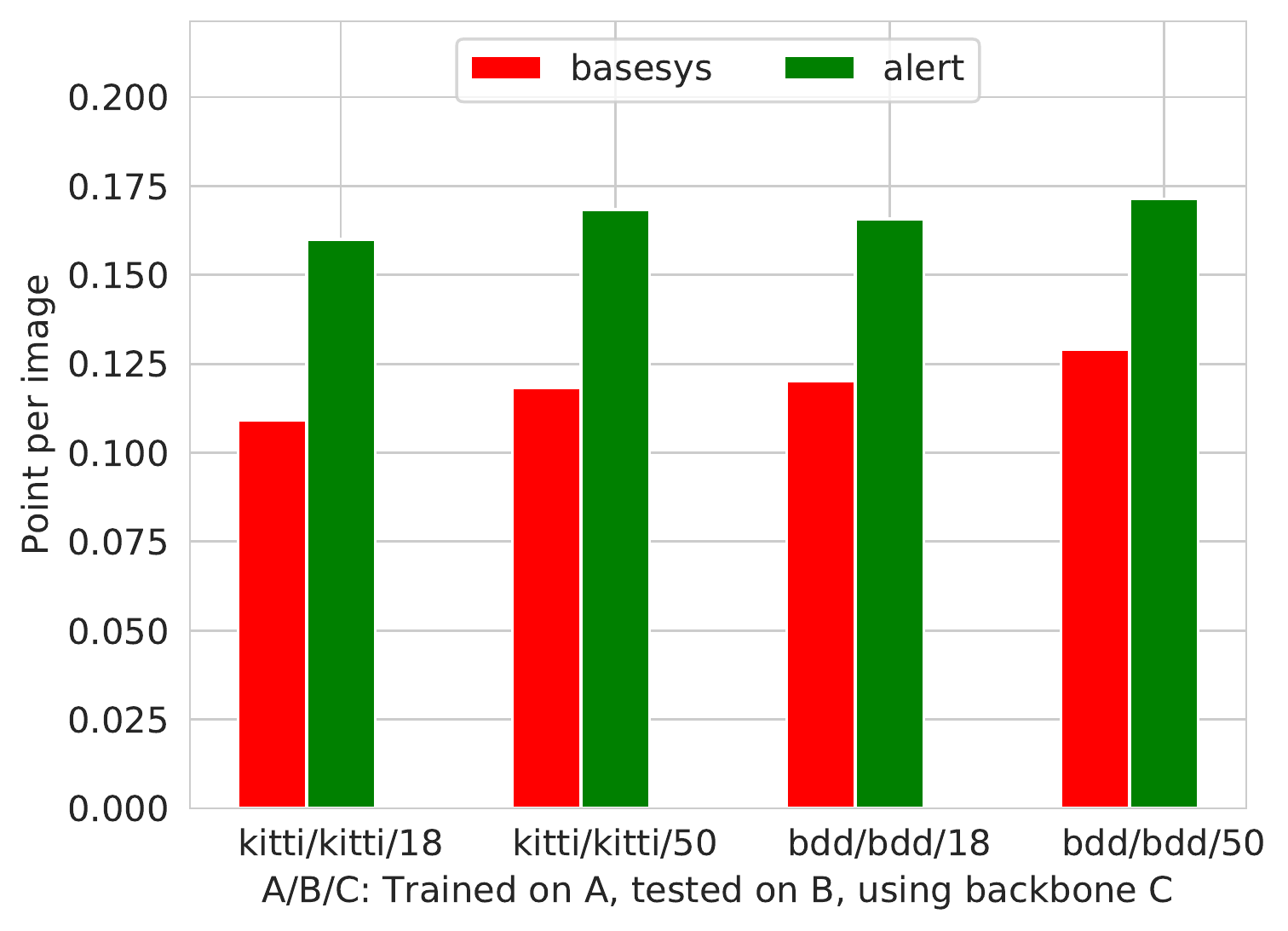}\label{fig:point_per_image_similar_dataset}} 
\subfloat[][]{\includegraphics[width=0.49\columnwidth]{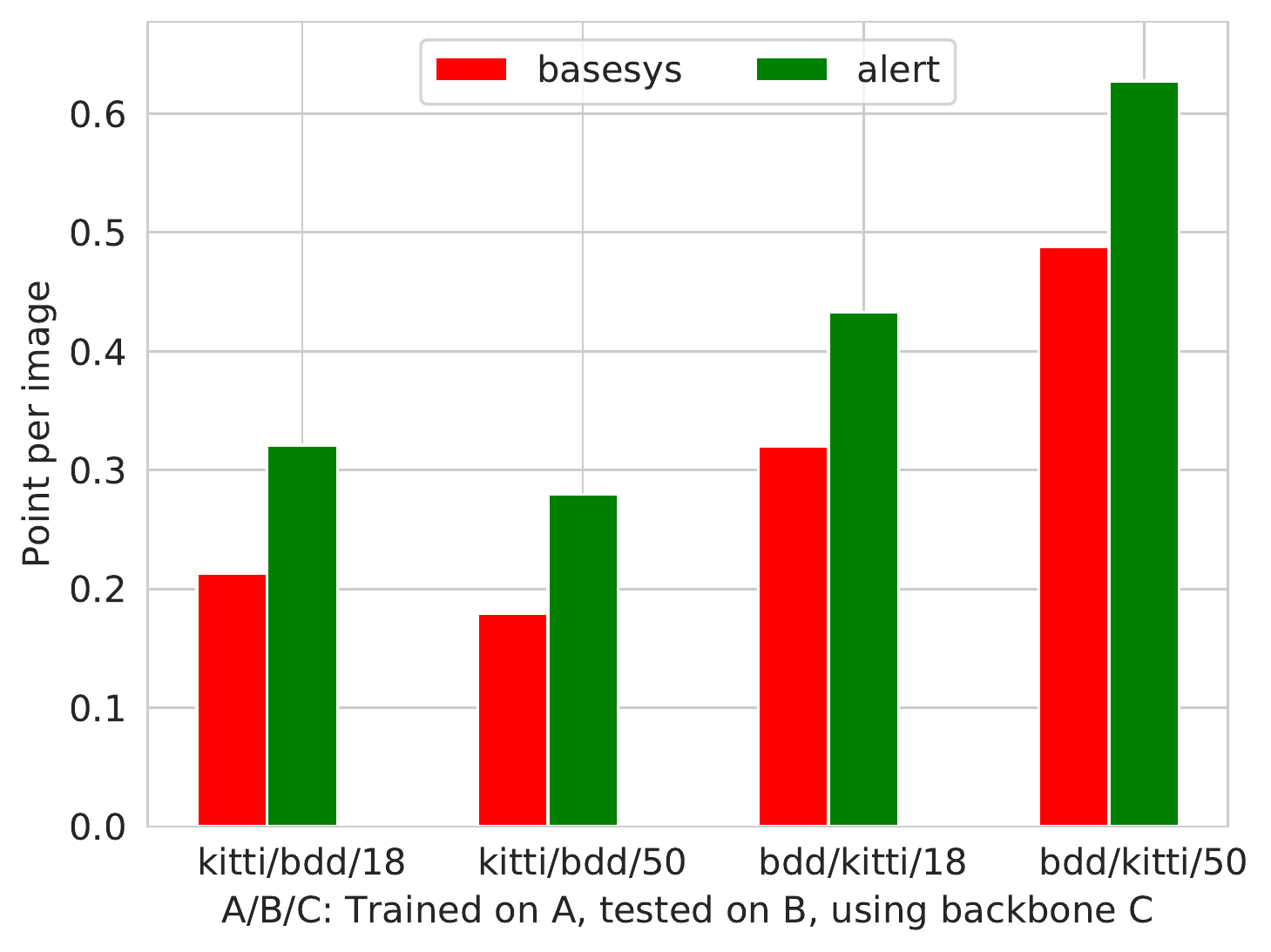}\label{fig:point_per_image_cross_dataset}} 
\caption{Risk-Averse Metric for the proposed \textit{alert} system. (a) Point per image earned by \textit{basesys} with and without considering \textit{alert} warning when trained and tested on similar dataset. (b) Point per image for \textit{basesys} with and without \textit{alert} system when trained and tested on different dataset. In both cases, \textit{basesys} earns more point per image when associated with \textit{alert}.}
\end{figure}

\subsection{Risk-Averse Metric}
In Risk-Averse Metric (RAM) \cite{zhang2014predicting} we evaluate \textit{alert}'s capability to trade-off the risk of making an incorrect decision with not making a decision at all. RAM gives \textit{basesys} $+1.0$ and $-0.5$ respectively for a correct and incorrect prediction. \textit{basesys} will get $0$ point if it does not make any decision considering the warning raised by \textit{alert}. For crucial system like self-driving car's object detection we expect \textit{basesys} to trade-off incorrect decision for no decision. In such case, \textit{basesys} can handover its control to some more competent systems. Figure~\ref{fig:point_per_image_similar_dataset} shows the point per image earned by \textit{basesys} when it operates with and without considering the warning raised by \textit{alert} in similar dataset settings. In all cases, \textit{basesys} earns more point per image if it abstains from making an incorrect decision taking \textit{alert}'s warning in consideration. Figure~\ref{fig:point_per_image_cross_dataset} shows the RAM metric for corss dataset settings. 

\begin{figure}[h!]
\centering
\subfloat[][]{\includegraphics[width=0.49\columnwidth]{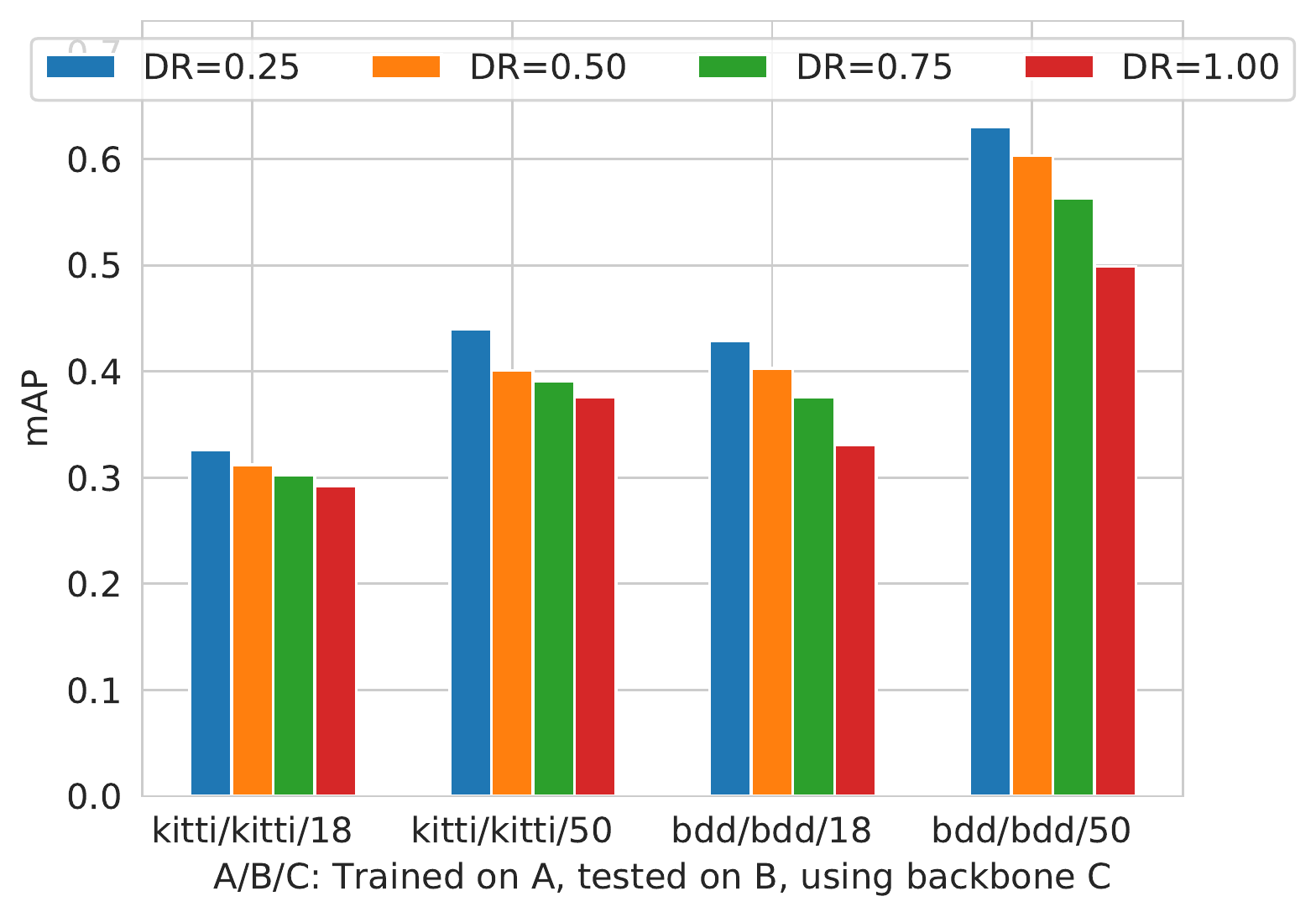}    \label{fig:dr_map_similar}}
\subfloat[][]{\includegraphics[width=0.49\columnwidth]{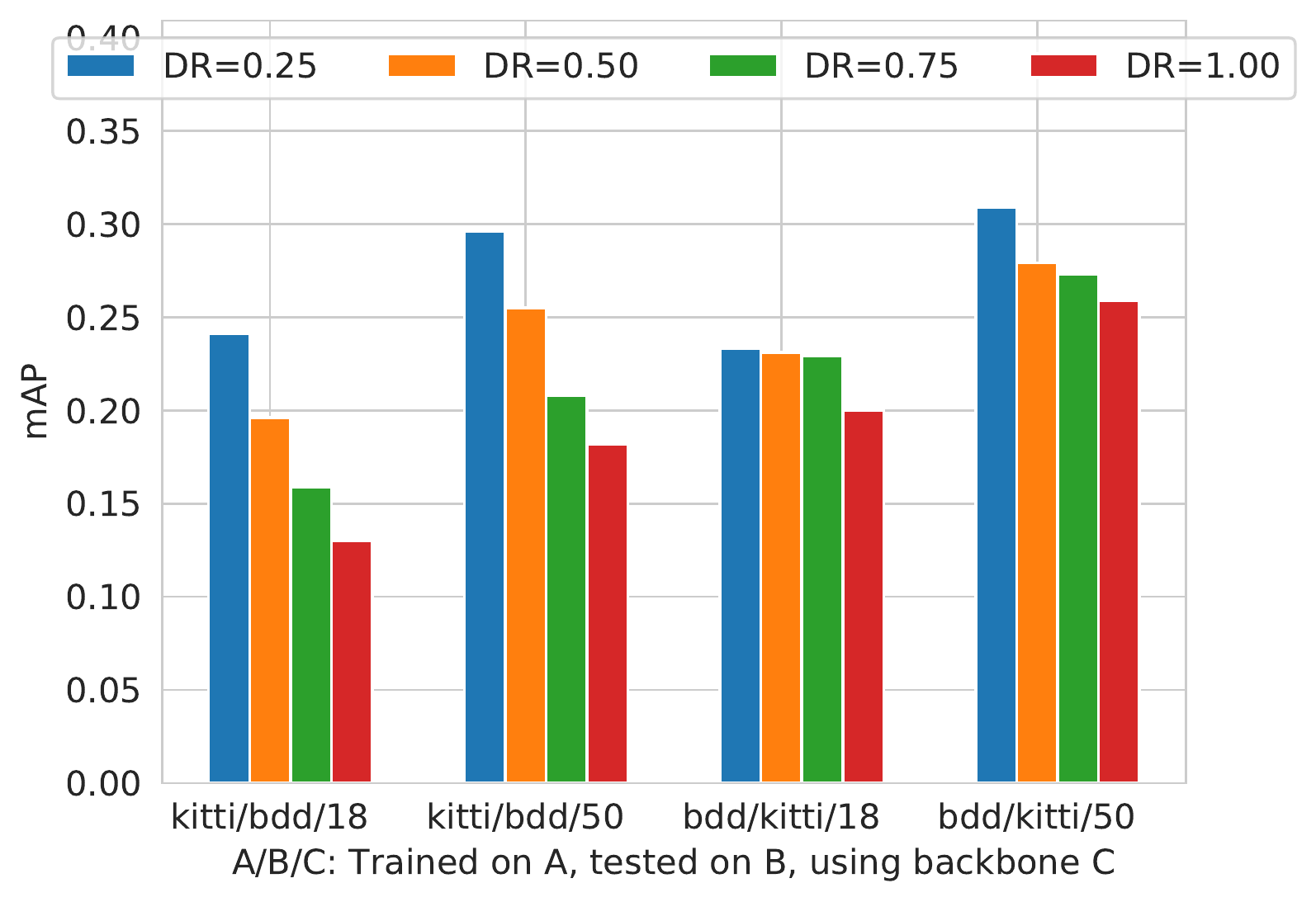}\label{fig:dr_map_corss}}  

\caption{mAP vs declaration rate metric for the proposed \textit{alert} system. We used four different declaration rate to calculate the corresponding mAP metric. An increasing declaration rate shows a gradual drop in the mean average precision. (a) shows the mAP vs DR metric for the similar dataset settings (b) mAP vs DR metric for cross dataset settings.}
\end{figure}

\subsection{mAP vs Declaration Rate Metric}
In this section we evaluate the \textit{basesys} accuracy score for different declaration rate (DR) \cite{zhang2014predicting}. Here, declaration rate is the proportion of images on which \textit{basesys} operates. The rest of the images are discard assuming that \textit{basesys}  per-frame mAP will be lower than the critical threshold on those images. To calculate this metric we first sort the images in the ascending order of \textit{alert} confidence. Next, mAP of top DR percentage of images are computed to plot the mAP vs DR metric. For a perfect \textit{alert} the mAP for low DR images would be very high and decrease gracefully as DR approaches to $1.0$. In Figure~\ref{fig:dr_map_similar} we show the mAP score for four different declaration rate in similar dataset settings. The mAP score drops gradually with the increasing declaration rate. Figure~\ref{fig:dr_map_corss} shows the mAP vs DR metric for cross dataset settings. In both cases, we use \textit{mean\_max\_std} features in \textit{alert} to identify \textit{basesys} performance drop. 

\section{Conclusion}
\label{sec:conclusion}
Deep learning-based object detection is a critical component of a wide variety of robotic applications, from autonomous vehicle to warehouse automation due to its accuracy and efficiency. However, its performance is a function of the deployment conditions and could drop below a critical threshold leading to increased risk. Although there is always room to improve accuracy and speed, safety is still a significant concern that should not be overlooked. To this end, we presented an introspection approach to performance monitoring of deep learning based object detection. We showed that our approach can improve safety by raising an alarm when per-frame mean average precision is detected to drop below a critical. We also showed that internal features from the detector itself could be used to predict when per-frame mAP degrade. Our results showed quantitatively the ability of our method to reduce risk by trading off making an incorrect detection with raising the alarm and absenting from detection.

\bibliographystyle{IEEEtran}
\bibliography{ref}
\end{document}